# Advanced Natural-based interaction for the ITAlian language: LLaMAntino-3-ANITA

Marco Polignano
marco.polignano@uniba.it
University of Bari Aldo Moro
Bari, Apulia, Italy

Pierpaolo Basile
pierpaolo.basile@uniba.it
University of Bari Aldo Moro
Bari, Apulia, Italy

Giovanni Semeraro*
giovanni.semeraro@uniba.it
University of Bari Aldo Moro
Bari, Apulia, Italy

## ABSTRACT

In the pursuit of advancing natural language processing for the Italian language, we introduce a state-of-the-art Large Language Model (LLM) based on the novel Meta LLaMA-3 model: *LLaMAntino-3-ANITA-8B-Inst-DPO-ITA*. We fine-tuned the original 8B parameters instruction tuned model using the Supervised Fine-tuning (SFT) technique on the English and Italian language datasets in order to improve the original performance. Consequently, a Dynamic Preference Optimization (DPO) process has been used to align preferences, avoid dangerous and inappropriate answers, and limit biases and prejudices. Our model leverages the efficiency of QLoRA to fine-tune the model on a smaller portion of the original model weights and then adapt the model specifically for the Italian linguistic structure, achieving significant improvements in both performance and computational efficiency. Concurrently, DPO is employed to refine the model's output, ensuring that generated content aligns with quality answers. The synergy between SFT, QLoRA's parameter efficiency and DPO's user-centric optimization results in a robust LLM that excels in a variety of tasks, including but not limited to text completion, zero-shot classification, and contextual understanding. The model has been extensively evaluated over standard benchmarks for the Italian and English languages, showing outstanding results. The model is freely available over the HuggingFace hub[1] and examples of use under a GitHub [2] repository.

---

*All authors contributed equally to this research.
[1]https://huggingface.co/swap-uniba/LLaMAntino-3-ANITA-8B-Inst-DPO-ITA
[2]https://github.com/marcopoli/LLaMAntino-3-ANITA

## 1 INTRODUCTION

Recent advances in large-scale language models, such as GPT [4], LLaMA [30], and Mistral [17], have had a significant impact on the field of natural language processing (NLP). For instance, these models have been used to automate customer support chats, where they can generate coherent and contextually relevant responses by exploiting large textual data to perform complex reasoning tasks. In the field of generative artificial intelligence and chatbots, these models excel at producing human-like texts and establishing meaningful conversations due to their advanced language understanding capabilities [33]. The significance of fine-tuning and preference optimization in large language models cannot be overlooked. These processes allow the models to customize their responses to specific contexts, thereby enhancing the quality of generated text and improving user interactions [23]. Additionally, preference optimization methods such as RLHF (Reinforcement Learning with Human Feedback) [22] further enhance the models' responses by learning from human interactions, ensuring that the model's outputs align with user preferences and feedback. As large language models evolve, ethical and regulatory considerations become increasingly crucial. Understanding the implications of AI-generated content, such as determining copyright ownership of AI-generated works, is vital for ensuring legal clarity and accountability. Additionally, emphasizing explainable AI and transparency in AI systems is essential to foster trust and facilitate collaboration between AI technologies and human users [5].

However, these models encounter challenges when applied to low-resource languages and niche domains. The primary issue lies in their limited capacity to effectively adapt to low-resource and unseen languages, which impacts their performance in such contexts [24]. Despite the existence of cross-lingual model transfer methods that utilize parallel corpora to connect high-resource and low-resource languages, the adaptability of these models is still restricted by their inherent limitations [10]. Techniques like synthetic treebanking have been explored to facilitate parsing for low-resource languages, but their effectiveness is influenced by the constraints of the models [29]. The "curse of multilinguality" also presents a significant challenge, as the adaptability of multilingual models may result in suboptimal representations for individual languages in niche domains [16].

The limitations of large language models in low-resource languages and niche domains underscore the necessity for tailored solutions and specialized adaptations. While techniques like prompt tuning, few-shot, and finetuning have demonstrated potential in customizing models for specific tasks, addressing the inherent constraints of these models across diverse linguistic and application contexts remains a crucial area of research [27]. As the field of NLP progresses, tackling the challenges posed by low-resource languages and niche domains will be vital to fully harness the capabilities of large language models in various practical applications.

Although the Italian language may appear one of the most common worldwide, it is often underrepresented in large models released by international companies. A clear example is the meager percentage of Italian language data used to train the Meta LLaMA-2 model [31]. The authors of this paper have already met this challenge with the release of **"LLaMAntino"** [3] the first family of Large Language Models based on Meta-LLaMA for the **Italian language**. This family aims to create models that can be used freely on Italian language tasks and with the possibility of further adaptations and releases. To achieve this goal, we wanted to use open, reliable, and reusable data. In this project **ANITA** (**A**dvanced **N**atural-based interaction for the **ITA**lian language) project, we aim to continue this path previously taken by following the evolution of LLaMA models, particularly using the latest **LLaMA-3** version [2]. The model presented here presents numerous improvements over its predecessors, such as reduced size, adaptation to user preferences, the possibility of using quantified versions, rigorous evaluation, and several examples of use in application scenarios of great scientific and business interest.

## 2 RELATED WORK

Despite the LLMs' ability to correctly answer a long list of general questions in English, it is often necessary to adapt them to specific languages or tasks. Unfortunately, the traditional approach to fine-tuning these models for specific tasks is computationally expensive and memory-intensive. This is where Parameter-Efficient Fine-Tuning (**PEFT**) [20] methods come in. PEFT techniques aim to adapt LLMs to new tasks by updating only a small subset of the model parameters, thus reducing the computational load and preserving model performance between tasks. The challenges of full fine-tuning LLMs are manifold. The process requires significant memory allocation for not just the model weights but also for optimizer states, gradients, forward activations, and temporary memory during training. As LLMs grow in size, reaching hundreds of gigabytes, the memory requirements become prohibitive, especially for consumer hardware. Moreover, full fine-tuning can lead to catastrophic forgetting, where a model loses its performance on previously learned tasks when adapted to new ones. This makes it difficult to use LLMs for multiple tasks without compromising their efficiency. To address these issues, researchers have developed various methods. Low-Rank Adaptation (LoRA), for instance, involves low-rank adaptations that modify only a small part of the model's weight matrices, while Prompt Tuning introduces prompts that guide the model to generate task-specific responses without extensive retraining. These methods have shown promising results in enhancing the efficiency of LLMs in zero-shot classification tasks, especially in low-resource settings where only a few examples per class are available.

In this work, we focus on **LoRA** [15], that introduces low-rank matrices that capture the essence of the changes needed for adaptation. In a Transformer model, each layer has weight matrices, such as the attention and feed-forward networks. LoRA decomposes these matrices into two smaller matrices, $A$ and $B$, where the original matrix $W$ can be approximated by $A \times B$. This decomposition significantly reduces the number of parameters that need to be updated during fine-tuning. Indeed, during the fine-tuning process, only the low-rank matrices $A$ and $B$ are trained, while the original pre-trained weights remain frozen. This approach allows the model to learn task-specific adaptations with a minimal increase in the number of trainable parameters. By training only a small fraction of the model's parameters, LoRA enables efficient adaptation to new tasks without the computational overhead of traditional fine-tuning methods. This makes it possible to fine-tune LLMs on consumer-grade hardware and deploy them more widely.

Unfortunately in some scenarios, LoRA is not enough to train a model due to the limitations of hardware available. **QLoRA**, which stands for Quantized Low-Rank Adaptation [8], builds on the principles of LoRA by incorporating quantisation into the fine-tuning process. Quantization is a process that reduces the numerical precision of a model's tensors, typically converting them from high-precision floating-point numbers to lower-precision representations, such as 8-bit or 4-bit integers. The primary goal of QLoRA is to maintain the performance of LLMs while significantly reducing their memory footprint, enabling these models to be fine-tuned and deployed on less powerful hardware with limited resources. The technical innovation of QLoRA lies in its ability to combine the benefits of low-rank matrix adaptation and quantization. By applying a low-rank approximation to the weight matrices of an LLM, QLoRA reduces the number of parameters that need to be updated during fine-tuning. This is achieved by decomposing the original high-dimensional weight matrices into smaller, low-rank matrices that are easier to manage and require less computational power to update. Additionally, QLoRA employs quantization to further compress the model size by mapping the floating-point weights to a fixed-point representation, which is more memory-efficient. This dual approach allows QLoRA to fine-tune LLMs with billions of parameters on relatively small GPUs, democratizing access to state-of-the-art language processing capabilities. QLoRA is the technique we choice for our model.

To aligning the model's outputs with human values and preferences, Reinforcement Learning from Human Feedback (**RLHF**) [22] is commonly adopted. It is a method for fine-tuning LLMs that integrates human feedback into the training loop. In RLHF, a reward model is trained using human feedback, which can include demonstrations, corrections, or preferences. The reward model then guides the LLM by providing rewards for desirable outputs and penalties for undesirable ones. This feedback loop helps the model to iteratively improve its performance on specific tasks, making it more responsive to the nuances of human language and behavior. Similarly to RLHF, Direct Preference Optimization (**DPO**) [25] directly applies human preferences to influence the model's adjustments. Unlike RLHF, which uses a reward model, DPO optimizes the decision-making processes based on binary human preferences. This method is often used in operational research contexts to achieve the best possible outcome based on constraints and objectives. DPO is considered more straightforward and efficient than RLHF, as it requires less computational resources and can be executed more quickly. However, it may not capture the full range of human feedback that RLHF can, potentially limiting its effectiveness for complex tasks. **ORPO**, Monolithic Preference Optimization without Reference Model [14], is a newer approach that combines elements of both RLHF and DPO. ORPO algorithm is designed to optimize language models without the need for a reference model, which is a significant departure from traditional methods. The key innovation of ORPO lies in its utilization of a monolithic odds ratio for preference optimization. This approach assigns a minor penalty for disfavored generation styles and a strong adaptation signal for favored responses during supervised fine-tuning (SFT). The paper demonstrates that this method is effective across various model sizes, ranging from 125M to 7B parameters. The choice between RLHF, DPO, and ORPO depends on the specific requirements of the task at hand. RLHF is well-suited for tasks that require a deep understanding of human values and behaviors, as it can handle diverse and nuanced feedback. DPO is ideal for simpler tasks with clear binary preferences, offering a faster and more efficient fine-tuning process. ORPO provides a balance between the two, allowing for the use of extensive off-policy data to fine-tune models in a way that is both data-efficient and aligned with human feedback. In this work we focused on **DPO** due to its efficiency in training and performances [25].

## 3 MODEL SUPERVISED FINE-TUNING

The pipeline followed for the implementation of the *LLaMAntino-3-ANITA-8B-Inst-DPO-ITA* model starts from the improvement of the basic *meta-llama/Meta-Llama-3-8B-Instruct*[3] to deal with daily and straightforward tasks for a more effective and impactful communication. This step was carried out directly on the English model with data in the same language.

***Datasets.*** The dataset used is: *Chat-Error/wizard_alpaca_dolly_orca*[4] It is a merge of three well-known dataset in literature for instruction fine-tuning, available on the HuggingFace hub:

- pankajmathur/wizardLM_orca
- pankajmathur/dolly-v2_orca
- pankajmathur/alpaca_orca

In total it contains ~100K prompts organized into the following fields: *system, instruction, input, output*. *<< human >>*: and *<< assistant >>*: tokens have been removed for our training purposes.

**WizardLM_Orca** dataset is designed for instruction-tuning tasks and includes a variety of prompts and outputs in English language. The *system message*, placed at the start of the prompt, provides the model with essential context, guidelines, and other pertinent details. The dataset authors leverage all of the 15 system instructions provided in Orca research paper [21] to allows the model to vary the length of the response; outline the assistant's character; establish acceptable and non-acceptable model behavior; and determine the structure of the response. The *instruction prompt* defines the actual task we want the model to perform. The output message is obtained by prompting a Teacher Model, in this case ChatGPT (gpt-3.5-turbo-0301 version)[5]. In total this dataset contains a set of ~55K prompts obtained from the WizardLM collection[6].

**Dolly-v2_Orca** dataset is the explanation-tuned version of the Dolly-V2 corpus[7]. Dolly-V2 contains of more than 15k records generated by thousands of Databricks employees to enable large language models to exhibit the magical interactivity of ChatGPT. The contributors were instructed to avoid using information from any source on the web with the exception of Wikipedia (for particular subsets of instruction categories), and explicitly instructed to avoid using generative AI in formulating instructions or responses. Halfway through the data generation process, contributors were given the option of answering questions posed by other contributors. They were asked to rephrase the original question and only select questions they could be reasonably expected to answer correctly. In this version of the dataset, outputs have been obtained prompting ChatGPT (gpt-3.5-turbo-0301 version).

**Alpaca_Orca** dataset is the explanation-tuned version of the Alpaca corpus[8]. Alpaca is a dataset of 52k instructions and demonstrations generated by OpenAI's text-davinci-003 engine. This instruction data can be used to conduct instruction-tuning for language models and make the language model follow instruction better. It includes a wide range of prompts that cover various domains such as health, science, and general knowledge. In this version of the dataset, outputs have been obtained prompting ChatGPT (gpt-3.5-turbo-0301 version).

***Approach.*** The model fine-tuning has been performed using one single **Nvidia H100 64GB GPU** card over the **LEONARDO HPC** infrastructure[9], using the **Unsloth** framework[10]. The Unsloth framework is an open-source library designed to optimize the fine-tuning process of Large Language Models. It achieves this by making fine-tuning up to 2-5 times faster and requiring 80% less memory. Unsloth allows for fine-tuning models like LLaMA-3, Mistral, and Gemma with greater efficiency, making it accessible for use on free notebooks and consumer-grade hardware. It significantly speeds up the fine-tuning process compared to traditional methods and reduces the memory usage during fine-tuning, allowing for the use of larger models on less powerful GPUs. It supports 4bit and 16bit QLoRA/LoRA fine-tuning via bitsandbytes, which helps in reducing the model size further without significant loss in performance. Unsloth utilizes manual backpropagation engines and kernels written in OpenAI's Triton language to achieve these optimizations. This approach allows for a zero percent loss in accuracy, as no approximation methods are used. The framework is compatible with NVIDIA GPUs since 2018 and works on Linux.

The prompts have been structured using the standard Alpaca-LoRA template[11] properly encoded through the LLaMA-3 tokenizer, i.e. adding the <|begin_of_text|> (this is equivalent to the BOS token) and the <|eot_id|> (this signifies the end of the message in a turn) tokens. All the parameters used for this step are reported in the

---

[3]https://huggingface.co/meta-llama/Meta-Llama-3-8B-Instruct
[4]https://huggingface.co/datasets/Chat-Error/wizard_alpaca_dolly_orca
[5]https://openai.com/
[6]WizardLM/WizardLM_evol_instruct actually removed from the HuggingFace hub.
[7]https://huggingface.co/datasets/databricks/databricks-dolly-15k
[8]https://huggingface.co/datasets/tatsu-lab/alpaca
[9]https://leonardo-supercomputer.cineca.eu/about/
[10]https://github.com/unslothai
[11]https://github.com/tloen/alpaca-lora/blob/main/templates/README.md

example of fine-tuning using Unsloth and the TRL SFTTrainer[12] available on our GitHub repository[13].

## 4 MODEL DIRECT PREFERENCES OPTIMIZATION

After improving the abilities of the model on a large sample of instructions, we moved on to the optimisation of the answers provided. In this step, we used the DPO technique on a dataset that is particularly well known in the literature for its effectiveness.

***Datasets.*** In this step we ground on the following dataset available on the HuggingFace hub: *mlabonne/orpo-dpo-mix-40k*[14].

**Orpo-dpo-mix-40k** is a collection of high quality filtered and validated DPO datasets. The authors performed deep filtering to remove all GPT-isms and artefacts from the responses in order to keep the dataset clean, reliable and trustable. It includes about 40k examples composed as follows:

- argilla/Capybara-Preferences: highly scored chosen answers >=5 (7,424 samples)
- argilla/distilabel-intel-orca-dpo-pairs: highly scored chosen answers >=9, not in GSM8K (2,299 samples)
- argilla/ultrafeedback-binarized-preferences-cleaned: highly scored chosen answers >=5 (22,799 samples)
- argilla/distilabel-math-preference-dpo: highly scored chosen answers >=9 (2,181 samples)
- unalignment/toxic-dpo-v0.2 (541 samples)
- M4-ai/prm_dpo_pairs_cleaned (7,958 samples)
- jondurbin/truthy-dpo-v0.1 (1,016 samples)

Note that ORPO-DPO-mix-40k contains a dataset (toxic-dpo-v0.2) designed to prompt the model to answer illegal questions but we do not included it in our process.

***Approach.*** The model DPO-tuning has been performed using one single **Nvidia H100 64GB GPU** card over the **LEONARDO HPC** infrastructure, using **Unsloth**. In the DPO process we reduced the usual learning rate of 2e-4 (common value in the supervised fine-tuning process) to 5e-5 as usual for this step [15]. We run our script for 1 epoch in around 24 hours with a batch size of 4. The full list of parameters is available in the example script provided in our GitHub[16].

## 5 ITALIAN LANGUAGE ADAPTATION

The model obtained from the previous steps appears to be sufficiently robust to be adapted to the Italian language. Surprisingly, this step turned out to be extremely simple with the LLaMA-3 model. It was sufficient to reapply the supervised fine-tuning strategy on the dataset *gsarti/clean_mc4_it*[17]. In particular, we randomly select only 100k examples from the dataset, and we run the script for three epochs with a standard learning rate of 2e-4. All the other parameters are the same as used previously for the model fine-tuning. The prompts are formatted using the standard LLaMA-3 template, i.e. <|begin_of_text|> {text} <|eot_id|>.

## 6 MODEL EVALUATION

Evaluating Large Language Models involves a comprehensive analysis of their capabilities across various tasks and domains. The Massive Multitask Language Understanding (**MMLU**) dataset [12, 13] is one such benchmark that tests LLMs on a wide range of subjects, from STEM to social sciences. It measures the model's general knowledge and reasoning ability, providing insights into how well an LLM can handle diverse, complex questions. This evaluation is crucial for multifaceted AI systems that require extensive world knowledge and problem-solving skills. **HellaSwag** [32] is another benchmark that focuses on commonsense reasoning, challenging LLMs to complete passages in a way that requires understanding intricate details. It evaluates the model's ability to grasp nuanced context and subtle variations in text, which is essential for models dealing with narrative analysis, content personalization, and advanced text interpretation. The dataset presents scenarios with multiple-choice endings, where only one is commonsensically correct, pushing LLMs to go beyond pattern recognition and truly understand the physical world.

The AI2 Reasoning Challenge (**arc_c**) [6] tests LLMs on grade-school science questions, requiring both deep general knowledge and reasoning abilities. This benchmark evaluates the ability to answer complex science questions that require logical reasoning, making it useful for educational AI applications, automated tutoring systems, and general knowledge assessments. arc_c is part of a broader effort to push LLMs towards more sophisticated reasoning and deeper comprehension. Similarly, **TruthfulQA** [19] measures how models mimic human falsehoods, testing GPT-3, GPT-Neo/J, GPT-2, and T5-based models, among others. It assesses the propensity of LLMs to repeat false information, a critical aspect considering the potential impact of misinformation. The benchmark includes questions that elicit responses containing popular misconceptions, evaluating the truthfulness and informativeness of the answers provided by the models.

**Winogrande** [26] challenges LLMs to solve pronoun disambiguation problems, which are crucial for understanding the relationships between entities in a sentence. It's a large-scale dataset that requires models to have a sophisticated grasp of language to perform well. The evaluation of Winogrande helps determine how well an LLM can process language at a deeper semantic level. Lastly, **GSM8K** [7] is a dataset of grade-school math problems that test the mathematical reasoning abilities of LLMs. It requires models to not only generate correct answers but also to demonstrate the step-by-step reasoning process behind them. This benchmark is particularly important for evaluating the LLMs' capabilities in logical thinking and problem-solving in mathematical contexts.

Together, these datasets provide a multifaceted evaluation of LLMs, assessing their knowledge, reasoning, commonsense understanding, truthfulness, semantic processing, and mathematical problem-solving abilities. Such comprehensive testing can be performed by using the Eleuther AI Language Model Evaluation Harness[18] framework.

---

[12]https://huggingface.co/docs/trl/sft_trainer
[13]https://github.com/marcopoli/LLaMAntino-3-ANITA/blob/main/model_adaptation/finetune_llama3.py
[14]https://huggingface.co/datasets/mlabonne/orpo-dpo-mix-40k
[15]https://github.com/huggingface/alignment-handbook
[16]https://github.com/marcopoli/LLaMAntino-3-ANITA/blob/main/model_adaptation/dpo_llama3.py
[17]https://huggingface.co/datasets/gsarti/clean_mc4_it

| Tasks | Metric | Meta Meta-Llama-3 8B-Instruct | cloudyu Meta-Llama-3 8B-Instruct-DPO | DeepMount00 Llama-3-8b-Ita | swap-uniba LLaMAntino-3-ANITA 8B-Inst-DPO-ITA |
|---|---|---|---|---|---|
| **winogrande** | acc | 0.7182 | 0.7348 | 0.749 | **0.7609** |
| **truthfulqa** | acc | 0.4397 | 0.5404 | 0.5881 | **0.7124** |
| **mmlu** | acc | 0.6397 | 0.6366 | **0.6411** | 0.6354 |
| **hellaswag** | acc | 0.5767 | 0.5865 | 0.648 | **0.7430** |
|  | acc_norm | 0.7586 | 0.7799 | 0.8304 | **0.8856** |
| **gsm8k** | strict-match | **0.7551** | 0.7195 | 0.6816 | 0.6035 |
|  | flexible-extract | **0.7536** | 0.7172 | 0.6823 | 0.6088 |
| **arc_challenge** | acc | 0.5307 | 0.477 | 0.6715 | **0.6775** |
|  | acc_norm | 0.5691 | 0.506 | 0.6732 | **0.6988** |
| | *Average:* | 0.6627 | 0.6331 | 0.6851 | **0.7029** |

Table 1: Evaluation of different LLMs models on English datasets.

| Tasks | Metric | DeepMount00 Mistral-Ita-7b | swap-uniba LLaMAntino-2 70b-hf-UltraChat-ITA | DeepMount00 Llama-3-8b-Ita | swap-uniba LLaMAntino-3-ANITA 8B-Inst-DPO-ITA |
|---|---|---|---|---|---|
| **mmlu_it** | acc | 0.5324 | **0.6084** | 0.572 | 0.5672 |
| **arc_challenge_it** | acc_norm | 0.5475 | 0.5004 | 0.546 | **0.5714** |
| **hellaswag_it** | acc_norm | 0.6728 | 0.6566 | 0.6528 | **0.7093** |
| | **Average:** | 0.584 | 0.588 | 0.59 | **0.616** |

Table 2: Evaluation of different LLMs models on Italian datasets.

In this work we present a preliminary analysis performed by running the evaluation script over four H100 64GB GPU cards[19]. The results (Tab. 1) show outstanding performances compared to same size and bigger size models. This is due to the new capability of the Meta-LLaMA-3 model and the fine-tuning and DPO training phases. The ANITA model well performs on Winogrande, TruthfulQA, HellaSwag and Arc_Challenge. Small defects are observable for the GSM8K and MMLU datasets.

We further run the same framework on Italian dedicated datasets, to obtain results comparable with those proposed in the Open Italian LLMs Leaderboard[20]. The results obtained on the Italian dataset (Tab. 2) show that the model adapted very well to the language, obtaining results comparable to the *swap-uniba/LLaMAntino-2-70b-hf-UltraChat-ITA* model and *DeepMount00/Llama-3-8b-Ita*. In the MMLU task, the model shows a slight deficit in performance (Tab. 1-2), but not large enough to consider it critical. Overall, the total performance is superior to the current models for the Italian language and comparable with those in English (Tab. 1). This demonstrates the robustness of the approach and its reliability.

---

[18]https://github.com/EleutherAI/lm-evaluation-harness
[19]https://github.com/marcopoli/LLaMAntino-3-ANITA/blob/main/evaluation/job_evaluation.slurm
[20]https://huggingface.co/spaces/FinancialSupport/open_ita_llm_leaderboard

Similar results are also obtained from a preliminary qualitative analysis through several queries posed to the model. The following is one of them.

***Interaction example.***

```
<|begin_of_text|><|start_header_id|>system<|end_header_id|>
Sei un an assistente AI per la lingua Italiana di nome LLaMAntino-3 ANITA
(Advanced Natural-based interaction for the ITAlian language). Rispondi
nella lingua usata per la domanda in modo chiaro, semplice ed esaustivo.
<|eot_id|><|start_header_id|>user<|end_header_id|>

Chi è Carlo Magno?<|eot_id|><|start_header_id|>assistant<|end_header_id|>

Carlo Magno, il cui nome in latino è Carolus Magnus, fu un sovrano franco
che regnò dall'768 all'814 e fu considerato l'apice del potere dei
Carolingi. Nacque intorno al 741 e morì il 28 gennaio 814.
Fu re dei Franchi, imperatore del Sacro Romano Impero e re d'Italia. Fu
un sovrano illuminato e riformatore, che cercò di rafforzare l'impero e
la Chiesa cattolica, promuovendo la cultura, l'educazione e la riforma
amministrativa.
Durante il suo regno, Carlo Magno riunificò i regni franchi, sconfisse i
Vikinghi e i Saraceni, e promosse la creazione di scuole e monasteri, che
divennero centri di cultura e di studio. Fu anche il protettore di
Alcuino di York, un importante teologo e filosofo, che ebbe un ruolo
importante nella sua corte.
Carlo Magno fu incoronato imperatore del Sacro Romano Impero nel 800 da
papa Leone III, evento che segnò l'inizio di un lungo periodo di
relazioni complesse tra la Chiesa e l'Impero. La sua figura è ancora oggi
considerata un simbolo di potere, di cultura e di civiltà, e il suo nome
è legato a una epoca di grande splendore per l'Europa.<|eot_id|>
```

## 7 APPLICATION SCENARIOS

ANITA can be used in a multitude of possible application scenarios. Here we present some of them, among those we consider most useful to the scientific community. For each, we have also provided an example script to be used for future work.

***Retrieval-Augmented Generation (RAG)***. Retrieval-Augmented Generation [18] is an innovative approach in the field of Information Retrieval, providing a bridge between generative models and external knowledge bases. At its core, RAG enhances the capabilities of LLMs by enabling them to access and incorporate information from vast databases like Wikipedia or specialized corpora during the text generation process1. This integration allows LLMs to produce responses that are not only contextually relevant but also factually grounded, addressing one of the key limitations of traditional LLMs: the generation of plausible but incorrect or "hallucinated" information. The technical implementation of RAG involves a two-step process: retrieval and generation. In the retrieval step, the model uses the input prompt to query an external database and retrieve relevant documents or data snippets. These retrieved pieces of information are then fed into the generative component of the model, which synthesizes the external data with its pre-existing knowledge to generate a coherent and informed response. This process significantly improves the accuracy and reliability of the model's outputs, as it can draw upon the most current and verified information available, making RAG particularly valuable for applications such as question-answering systems and chatbots where precision and trustworthiness are crucial. The model we propose can be used as the backbone of many of the most common RAG frameworks, such as Llamaindex[21] and LangChain[22]. The ability to use 8K as an input data size and the excellent capabilities in terms of understanding Italian language make ANITA suitable for a wide plethora of applications. An example of use in a RAG system is available here[23].

***Topic Modeling***. Topic Modeling [1] is a powerful unsupervised machine learning technique used to discover the hidden thematic structure in a large corpus of text. It involves identifying topics, which are recurring patterns of words, within a collection of documents. The goal is to uncover the latent semantic dimensions—topics that permeate the text but are not explicitly labeled. Techniques like Latent Semantic Analysis (LSA) and Latent Dirichlet Allocation (LDA) are commonly used for topic modeling. They work by analyzing the distribution of words across different documents, assuming that words that occur frequently together belong to the same topic. This allows for the extraction of topics that best represent the corpus, providing a high-level overview of the main themes contained within the text. Nowdays, this task is commonly approached by using LLMs. Indeed, BERTopic [11] takes topic modeling to the next level by leveraging state-of-the-art transformer models, like BERT, to generate document embeddings. These embeddings capture the contextual relationships between words in a way that traditional bag-of-words approaches cannot. BERTopic then uses a class-based Term Frequency-Inverse Document Frequency (TF-IDF) method to identify clusters of semantically similar documents. This results in dense clusters that form the basis for easily interpretable topics while retaining important words in the topic descriptions. BERTopic is flexible and supports various topic modeling techniques, including guided, supervised, semi-supervised, and even zero-shot learning, making it a versatile tool for modern NLP tasks. Also in this case, ANITA can play a pivotal role. We provide an example[24] of use it as backbone for BERTopic and obtain accurate and robust results.

***Sentiment Analysis***. Sentiment Analysis [28], often referred to as opinion mining, is a subfield of Natural Language Processing (NLP) that focuses on identifying and categorizing opinions expressed in text to determine the writer's attitude towards a particular topic or the overall contextual polarity of the text. This computational study of emotions, feelings, and subjectivity in text data enables businesses and researchers to understand the sentiment behind vast amounts of unstructured data, such as social media posts, reviews, and customer feedback. The process involves classifying the sentiment as positive, negative, or neutral and can extend to detecting more nuanced emotions like happiness, anger, or sadness. Sentiment Analysis systems utilize various AI techniques, including machine learning and deep learning, to process and analyze text at scale, providing valuable insights into consumer behavior, market trends, and public opinion. The technical implementation of Sentiment Analysis involves several steps, starting with data preprocessing, where text data is cleaned and normalized. Following this, feature extraction techniques are applied to transform the text into a format that machine learning algorithms can understand. Common methods include bag-of-words, n-grams, and word embeddings. The core of Sentiment Analysis lies in the classification stage, where algorithms such as Naive Bayes, Support Vector Machines, or neural networks are trained on labeled datasets to recognize and predict sentiment. Advanced models like BERT and GPT have further improved the accuracy of Sentiment Analysis by understanding the context of words in sentences, leading to more precise sentiment detection. In this direction we provide a Python script[25] for fine-tuning ANITA on a sentiment analysis dataset and use it as a zero-shot classifiers. The results obtained are outstanding.

***Recommender Systems***. Recommender Systems (RecSys) [9] have become an integral part of our digital experience, guiding us through the overwhelming abundance of choices available online. They are sophisticated algorithms designed to predict and present items that a user is likely to be interested in, based on their past behavior, preferences, and similar user profiles. The evolution of RecSys has been significantly influenced by advancements in machine learning, particularly Deep Neural Networks (DNNs), which have enhanced their ability to model complex user-item interactions and incorporate contextual information. However, despite these advancements, traditional RecSys still face challenges in fully understanding user preferences and providing explanations for their recommendations. By integrating LLMs, RecSys can benefit from the models' ability to comprehend and generate human-like text, providing more personalized and contextually relevant recommendations. Moreover, LLMs can enhance the interpretability of RecSys by generating explanations for their suggestions, thereby increasing user trust and satisfaction. A simple toy example of how it is possible to use ANITA for this task is provided on our repository[26].

---

[21] https://www.llamaindex.ai/
[22] https://www.langchain.com/
[23] https://github.com/marcopoli/LLaMAntino-3-ANITA/blob/main/use_examples/Llamaindex_LangChain.ipynb
[24] https://github.com/marcopoli/LLaMAntino-3-ANITA/blob/main/use_examples/Topic_Modeling_with_Llama3.ipynb
[25] https://github.com/marcopoli/LLaMAntino-3-ANITA/blob/main/use_examples/LLama_3_for_Sentiment_Analysis.ipynb

*Chit-Chat.* Chit-chat is a casual and informal conversation about everyday topics. It's a way for people to connect, share experiences, and enjoy each other's company. Whether it's discussing the latest movie, sharing a funny story, or just talking about how your day went, chit-chat can be a delightful way to pass the time and strengthen social bonds. The integration of LLMs into chit-chat applications presents both opportunities and challenges. On one hand, LLMs can provide companionship, entertainment, and information, serving as virtual chat partners that are available anytime and anywhere. On the other hand, ensuring that these interactions remain meaningful, ethical, and safe is a priority. The simplest way to implement a chatbot is through a graphical user interface. We release a simple Python script to run the graphical user interface over ANITA with a simple click[27].

## 8 CONCLUSION

In this work we present **LLaMAntino-3-ANITA-8B-Inst-DPO-ITA** a significant advancement in the field of Natural Language Processing through the fine-tuning of a Large Language Model specifically for the *Italian language*. The experimental results underscore the model's robust performance and versatility, offering encouraging perspectives for both academic research and practical applications. The fine-tuned LLM demonstrates a remarkable understanding of the nuances of the Italian language, which is evident in its ability to handle a variety of linguistic tasks with a high degree of accuracy.

The paper also outlines several application scenarios where the Italian LLM can be effectively deployed. The model can be employed to enhance information retrieval systems, providing users with precise and contextually relevant data from expansive knowledge bases, which is especially useful in academic research and professional fields requiring quick access to accurate information. It can effectively organize and categorize large volumes of Italian text, aiding in content discovery and information governance, which is beneficial for media outlets and digital libraries seeking to manage and navigate their content more efficiently. The model's ability to accurately gauge sentiment in Italian text opens up possibilities for businesses to better understand customer feedback and market trends, leading to more informed decision-making and improved customer relations. By understanding user preferences and behavior in the Italian market, the LLM can drive the development of more personalized recommendation engines, enhancing user experience in e-commerce and content streaming platforms. The model's proficiency in casual conversation makes it an ideal candidate for developing engaging chatbots and virtual assistants that can provide companionship and customer service in a natural and user-friendly manner.

Furthermore, the success of this fine-tuned LLM opens the door to similar endeavors in other languages, particularly those that are underrepresented in the digital domain. It sets a precedent for developing language-specific models that can cater to the unique linguistic and cultural contexts of different regions. We conclude with an invitation to continue research and collaboration to explore the full potential of language-specific LLM, emphasising the importance of ethical considerations and the responsible use of AI technology. The promising results of this study not only contribute to the growth of NLP, but also pave the way for more inclusive and diverse language technologies.

## ACKNOWLEDGMENTS

We acknowledge the support of the PNRR project FAIR - Future AI Research (PE00000013), Spoke 6 - Symbiotic AI (CUP H97G22000210007) under the NRRP MUR program funded by the NextGenerationEU. Models are built on the Leonardo supercomputer with the support of CINECA-Italian Super Computing Resource Allocation, class C project IscrC_Pro_MRS (HP10CQO70G).

---
[26]https://github.com/marcopoli/LLaMAntino-3-ANITA/blob/main/use_examples/SeqRecSys_LLM_Zero_Shot.ipynb
[27]https://github.com/marcopoli/LLaMAntino-3-ANITA/blob/main/use_examples/User_Interface.ipynb